\documentclass{article}

\usepackage[preprint]{neurips_2024}

\usepackage[utf8]{inputenc} 
\usepackage[T1]{fontenc}    
\usepackage[hyperfootnotes=false]{hyperref}       
\usepackage{url}            
\usepackage{booktabs}       
\usepackage{nicefrac}       
\usepackage{microtype}      
\usepackage{xcolor}         
\usepackage{amsmath}
\usepackage[T1]{fontenc}
\usepackage{amsfonts} 
\usepackage{multirow}
\usepackage{colortbl}
\usepackage{wrapfig}
\usepackage{tcolorbox}
\usepackage{tablefootnote}
\usepackage{threeparttable}
\usepackage[symbol]{footmisc}

\definecolor{cyan}{cmyk}{.3,0,0,0}

\newtoggle{comment}
\toggletrue{comment}

\newcommand{\VarSty}[1]{\textnormal{\ttfamily\color{cyan!90!black}#1}\unskip}

\newcommand{\modelname}{StreamAdapter}
\title{StreamAdapter: Efficient Test Time Adaptation from Contextual Streams}

\author{
  Dilxat Muhtar\textsuperscript{\rm 1,*,$\dagger$}
  Yelong Shen\textsuperscript{\rm 2,*},
  Yaming Yang\textsuperscript{\rm 2},
  Xiaodong Liu\textsuperscript{\rm 2},
  Yadong Lu\textsuperscript{\rm 2},
  \\
  \textbf{
  Jianfeng Liu\textsuperscript{\rm 2},
  Yuefeng Zhan\textsuperscript{\rm 2},
  Hao Sun\textsuperscript{\rm 2},
  Weiwei Deng\textsuperscript{\rm 2},
  Feng Sun\textsuperscript{\rm 2},}
  \\
  \textbf{
  Xueliang Zhang\textsuperscript{\rm 1},
  Jianfeng Gao\textsuperscript{\rm 2},
  Weizhu Chen\textsuperscript{\rm 2},
  Qi Zhang\textsuperscript{\rm 2}
  }
  \\
  \textsuperscript{\rm 1} 
  Nanjing University
  \qquad
  \textsuperscript{\rm 2} Microsoft
  \\
  dmuhtar@smail.nju.edu.cn,
  zxl@nju.edu.cn
  \\
\{yelong.shen, xiaodl, yadonglu, jianfengliu, yuefeng.zhan\}@microsoft.com\\
\{hasun, dedeng, jfgao, wzchen, qizhang\}@microsoft.com
}

\begin{document}

\maketitle

\begin{abstract}
\footnotetext{* Equal Contribution}
\footnotetext{{$^{\dagger}$} Work done during internship at Microsoft}
In-context learning (ICL) allows large language models (LLMs) to adapt to new tasks directly from the given demonstrations without requiring gradient updates.
While recent advances have expanded context windows to accommodate more demonstrations, this approach increases inference costs without necessarily improving performance.
To mitigate these issues, We propose \modelname, a novel approach that directly updates model parameters from context at test time, eliminating the need for explicit in-context demonstrations.
\modelname~employs context mapping and weight absorption mechanisms to dynamically transform ICL demonstrations into parameter updates with minimal additional parameters.
By reducing reliance on numerous in-context examples, \modelname~significantly reduce inference costs and allows for efficient inference with constant time complexity, regardless of demonstration count.
Extensive experiments across diverse tasks and model architectures demonstrate that \modelname~achieves comparable or superior adaptation capability to ICL while requiring significantly fewer demonstrations.
The superior task adaptation and context encoding capabilities of \modelname~on both language understanding and generation tasks provides a new perspective for adapting LLMs at test time using context, allowing for more efficient adaptation across scenarios and more cost-effective inference.

\end{abstract}
\section{Introduction}
Large language models (LLMs) have emerged as a powerful tool in natural language processing, demonstrating exceptional performance across a diverse range of tasks, 
including text generation~\citep{Yuan2022WordcraftSW}, question answering~\citep{Kumar2023ConformalPW}, open-ended conversations~\citep{Zhang2023SpeechGPTEL}, and mathematical problem-solving~\citep{Shao2024DeepSeekMathPT}.
A key factor behind the success of LLMs is their ability to perform in-context learning (ICL)~\citep{brown2020languagemodelsfewshotlearners}, where the model adapts to new tasks by conditioning on a small number of input-output demonstrations provided in the context.
Without any gradient updates, ICL enables LLMs to acquire new knowledge and capabilities at test time, while also enabling LLMs to solve complex tasks through step-by-step guidance \citep{wei2023chainofthoughtpromptingelicitsreasoning}.

Despite its remarkable capabilities, ICL faces several limitations that hinder its full potential. 
Firstly, the effectiveness of ICL heavily depends on the quality and relevance of the provided demonstrations, making the selection of appropriate examples a challenging task that often requires domain expertise~\citep{Agarwal2024ManyShotIL,sahoo2024systematicsurveypromptengineering}. 
Moreover, the number of demonstrations that can be included is constrained by the model's context window size. While recent advancements have expanded these windows \citep{Ding2024LongRoPEEL,Reid2024Gemini1U}, accommodating more examples introduces significant computational overhead \citep{Fu2024ChallengesID}.

Although recent studies have attempted to use heuristic rules to select the most important subset of context to improve the robustness and efficiency of ICL \citep{li2024snapkvllmknowslooking,zhang2023h2oheavyhitteroracleefficient}, these methods inevitably cannot ensure that the discarded tokens that are currently unimportant will not become important in future decoding steps. Other investigations have focused on constructing meta-ICL approaches to enhance ICL's robustness and reduce reliance on perfect prompts \citep{codaforno2023metaincontextlearninglargelanguage}. Yet, these methods remain constrained by limited context length and often require hand-crafted prompt strategies, potentially leading to suboptimal performance.
On the other hand, recent studies suggest that ICL is actually performing a meta-gradient update for adapting to new tasks given the context information \citep{Dai2023WhyCG,Oswald2022TransformersLI}. 
These findings lead us to a crucial question: \textit{Instead of implicitly "updating" model parameters to adapt to a new domain or task via context, is it possible to directly convert the context into parameter updates, thus updating the network at test time without any backpropagation and without requiring demonstrations in the context window?}

To answer this question, we propose \modelname, a novel approach that leverages the inherent capabilities of LLMs to encode context information into their parameters. 
Instead of storing demonstrations explicitly in the input context, \modelname~dynamically maps these demonstrations into temporary parameter updates. 
This approach allows the model to benefit from context to adapt to new tasks similar to ICL at test-time, without consuming the context window or requiring backpropagation, thereby reducing the resource requirements of traditional ICL methods. 
\modelname~employs two key mechanisms to achieve this goal:
a) Context Mapping: This mechanism utilizes intra-chunk cross-attention and inter-chunk recurrence to adaptively condense the variable cached context into a constant context state for each parameter in the linear layer of LLMs.
b) Weight Absorption: The condensed context state interacts with two lightweight low-rank matrices to be absorbed into the original model parameters. This process updates the LLM's knowledge with minimal additional learnable parameters and incurs no additional inference latency.
By combining these mechanisms, \modelname~effectively distills the context into parameter updates, allowing for more efficient test-time adaptation (TTA).
Comprehensive experiments across diverse language understanding and long-context generation tasks, with various model architectures and scales, demonstrate that \modelname~achieves comparable or superior adaptation capability to full context evaluation while outperforming other context compression variants and TTA methods.
Moreover, \modelname~not only demonstrates constant inference generation time and lower memory consumption compared to full context generation, but also shows better scalability when provided with more adaptation context and improved robustness across various scenarios.

The contributions of our work can be summarized as follows:
\begin{itemize}
    \item We propose a new TTA strategy, \modelname, that directly maps the given context into parameter updates, rather than conditioning on the context. 
    This method enables models to quickly adapt to new tasks or acquire new temporary knowledge at test time like ICL, but with fewer or no demonstrations in context, thereby reducing memory consumption and inference time.
    \item We design \modelname~with innovative context mapping and low-rank adaptation mechanisms. 
    These allow \modelname~to map the context into parameter updates with minimal additional learning parameters and without inducing any additional inference latency.
    \item  We validate \modelname~on both language understanding and language generation tasks across various model scales and architectures. The results demonstrate the effectiveness of \modelname~over ICL and other TTA methods in various adaptation scenarios. Analyses of efficiency and robustness further highlight \modelname's advantages in terms of computational resources and generalization capabilities.
\end{itemize}
\section{Related Work}
\subsection{In-Context Learning}
ICL enables LLMs to acquire new knowledge or adapt to new tasks using in-context examples at test time without any gradient updates \citep{brown2020languagemodelsfewshotlearners}. 
Recent studies show that with proper instruction and more demonstrations, ICL can surpass model fine-tuning and mitigate inherent biases in pre-trained LLMs 
\citep{Agarwal2024ManyShotIL,li2024longcontextllmsstrugglelong}.
This exceptional capability has inspired research into ICL's working mechanisms, leading to various hypotheses such as induction heads \citep{olsson2022incontextlearninginductionheads}, task vectors \citep{hendel2023incontextlearningcreatestask,zheng2024distributedrulevectorskey}, and structured task hypothesis \citep{Li2024WhatDL}. 
A popular assumption posits that ICL performs meta-gradient descent during inference. \citet{Oswald2022TransformersLI} demonstrate how a linear attention-only transformer model can implicitly perform a gradient descent-like procedure, while \citet{Dai2023WhyCG} compare standard gradient descent-based fine-tuning and ICL, revealing that transformer attention in ICL exhibits a dual form of gradient descent-based optimization.
Inspired by these findings, our work seeks to develop a learning algorithm that directly performs parameter updates from the context without backpropagation at test time, aiming to achieve performance similar to ICL while requiring limited or no demonstrations in the context.

\subsection{Test-Time Adaptation}
Test-time adaptation (TTA) enhances model capabilities at inference by learning directly from test data \citep{niu2024test}. 
In-context learning (ICL) represents a form of TTA where models adapt to new tasks using demonstrations within the context at test time. Recent TTA research primarily follows two directions:
a) Condition Augmentation:
This approach focuses on modifying the context conditioning to improve performance, either through heuristic rules for adjusting conditional prediction distributions \citep{li2024snapkvllmknowslooking,zhang2023h2oheavyhitteroracleefficient} or through sampling strategies like best-of-N and reward-model based sampling \citep{cobbe2021training,chen2024alphamath,yao2024tree}.
b) Parameter Updates: 
This direction explores modifying model parameters at inference time. Early approaches build on fast weight programming \citep{hinton1987using}, exemplified by fast weight programmers \citep{Schlag2021LinearTA} and Hopfield networks \citep{Ramsauer2020HopfieldNI}, which update pre-trained weights using input-based products.
Meta-learning approaches \citep{finn2017model,beck2023hypernetworks} employ hypernetworks to generate auxiliary parameters for test-time adaptation. TempLoRA \citep{wang2024greatertextcomesgreater} extends this concept by training chunk-specific low-rank adapters \citep{hu2021lora} for next-chunk prediction. Recent work \citep{sun2024learning} formalizes test-time parameter updates through self-supervised learning with TTT-Linear and TTT-MLP, treating model parameters as latent RNN states.

Our approach, \modelname, aligns with parameter update methods but uniquely maps context directly into parameter updates at test time without backpropagation.

\subsection{Low-Rank Adaptation}
Inspired by the observation that pre-trained models have low intrinsic dimension during fine-tuning \citep{aghajanyan2020intrinsic}, low-rank adaptation (LoRA) \citep{hu2021lora} employs two trainable low-rank matrices to estimate the accumulated gradient updates, thereby adapting pre-trained models with minimal additional parameters.
Given its lower inference latency and superior adaptation performance, LoRA has been widely adopted, with subsequent research enhancing its efficiency and stability through dynamic rank allocation across layers \citep{zhang2023adalora} and further matrix decomposition \citep{liu2024dora}.
Our work also employs low-rank adaptation to adapt LLMs with minimal parameters. However, instead of training the adapter for specific tasks or datasets, \modelname~learns directly from previous context at test time, enabling more customized and flexible adaptation.

\section{Method}
\label{sec:method}
\begin{figure}[t!]
  \centering
  \includegraphics[width=\linewidth]{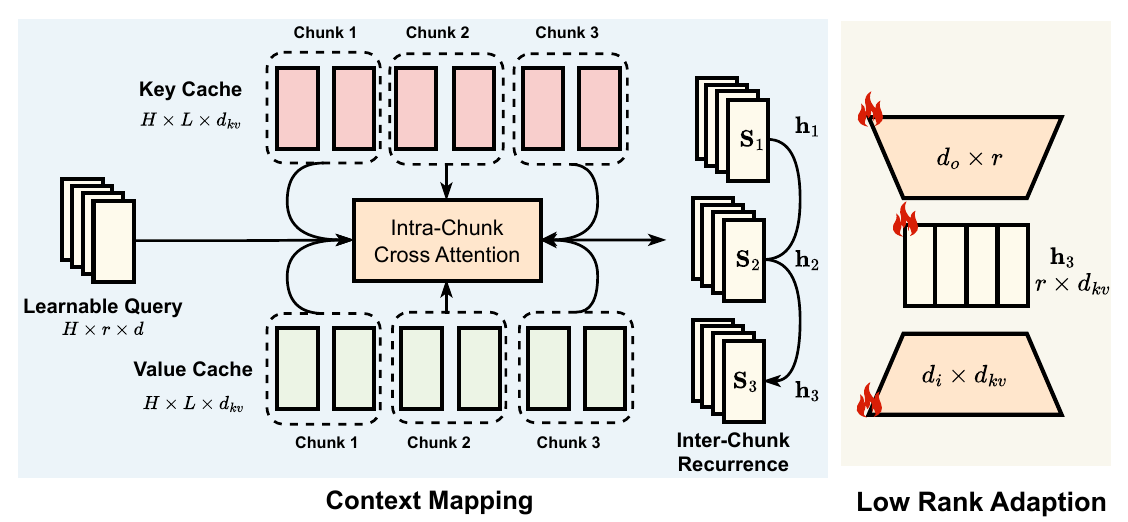}
    \caption{Overall structure of \modelname. \modelname~maps the KV cache into a context state using intra-chunk cross-attention and inter-chunk recurrence, then connects two low-rank matrices through the context state to update the model parameters for absorbing context information into model weights}
  \label{fig:architecture}
\end{figure}

We propose \modelname~to directly map contextual information into parameter updates, serving as a temporary weight-level associative memory that encodes new knowledge and adapts to new tasks without relying on full explicit context.
The overall structure of \modelname is presented in Figure~\ref{fig:architecture}.
\modelname~utilizes intra-chunk cross-attention and inter-chunk gated recurrence to adaptively map sparse context information into constant-sized context states (context mapping). 
These states are then absorbed into pre-trained weights through low-rank adaptation.

In the following subsections, we will briefly describe the duality between ICL and model parameter updates through gradient descent, shedding light on the fundamental motivation behind \modelname~and its formalization. 
We will then explore the details of \modelname~context mapping and weight absorption mechanisms.

\subsection{Duality between In-Context Learning and Weight Updates}
Recent studies have highlighted the inherent similarities between ICL and parameter updates through gradient descent~\citet{Dai2023WhyCG,Oswald2022TransformersLI}.
Specifically, let $x_i$ be the current input token, $\mathbf{X}'$ be the previous context, and $\mathbf{W}_q, \mathbf{W}_k, \mathbf{W}_v$ be the projection matrices of the self-attention (SA) layer.
By approximating standard SA with linear attention, the output of single-head SA is formulated as:
\begin{equation}
    \begin{aligned}
        \mathcal{F}_{\text{ICL}}(x_i) &\approx \mathbf{W}_v[\mathbf{X}', x_i](\mathbf{W}_k[\mathbf{X}', x_i])^T\mathbf{W}_qx_i \\
                          & = \mathbf{W}_vx_i(\mathbf{W}_kx_i)^T\mathbf{W}_qx_i + \mathbf{W}_v\mathbf{X}'(\mathbf{W}_k\mathbf{X}')^T\mathbf{W}_qx_i \\
                          & = (\mathbf{W}_\text{0} + \Delta\mathbf{W}_{\text{ICL}})\mathbf{W}_qx_i,
    \end{aligned}
\end{equation}
where $\mathbf{W}_\text{0}=\mathbf{W}_vx_i(\mathbf{W}_kx_i)^T$ are the initial result without any context, and $\Delta\mathbf{W}_{\text{ICL}}=\mathbf{W}_v\mathbf{X}'(\mathbf{W}_k\mathbf{X}')^T$ represents the "parameter updates" obtained from the given context.
Moreover, denoting $\Delta\mathbf{W}_k$ and $\Delta\mathbf{W}_v$ as the accumulated gradient updates from fine-tuning, the result of linear attention can be expressed as:
\begin{equation}
    \begin{aligned}
        \mathcal{F}_{\text{FT}}(x_i) &= (\mathbf{W}_v + \Delta\mathbf{W}_v)x_ix_i^T(\mathbf{W}_k + \Delta\mathbf{W}_k)\mathbf{W}_qx_i \\
        & = (\mathbf{W}_\text{0} + \Delta\mathbf{W}_{\text{FT}})\mathbf{W}_qx_i.
    \end{aligned}
\end{equation}
From the similarity between $\mathcal{F}_{\text{ICL}}(x_i)$ and $\mathcal{F}_{\text{FT}}(x_i)$, it can be hypothesized that ICL actually functions as a meta-optimizer, updating the underlying parameters through context-level associations \citep{Dai2023WhyCG}.

In this study, we delve deeper into the potential of leveraging context to directly update model parameters, thereby integrating context information into the model’s weights. 
The objective of \modelname~is to learn a mapping function $\mathcal{F}$ that, given context $\mathbf{X}'$, maps the key-value (KV) caches $\mathbf{W}_k\mathbf{X}'$ and $\mathbf{W}_v\mathbf{X}'$ of $\mathbf{X}'$ to parameter update $\Delta\mathbf{W}$:
\begin{equation}\label{eq:formalize}
    \mathcal{F}(\mathbf{W}_k\mathbf{X}', \mathbf{W}_v\mathbf{X}') \rightarrow \Delta\mathbf{W}.
\end{equation}
We anticipate that updating the model parameters with $\Delta\mathbf{W}$ will achieve results comparable to full ICL without the need for complete demonstrations filling the context window.

\subsection{Context Mapping}\label{sec:context_mapping}
The KV cache scales linearly with the context, whereas the model’s parameters remain constant in size. 
Consequently, a context mapping strategy that condenses the cache information into a fixed-size state is essential for absorbing context information into the model’s weights.
The most straightforward approach to achieving this is to compress the KV cache into a latent hidden state, similar to recurrent models~\citep{10.1162/neco.1997.9.8.1735,gu2023mamba}. 
However, token-by-token recurrence requires substantial memory, as it necessitates materializing all time step states. 
To mitigate this issue, we propose splitting the KV cache into fixed-size chunks and leveraging a number of learnable queries to summarize each chunk of caches. We then perform inter-chunk recurrence across each chunk of summarized results to convert the cache into a constant-size context state.
More specifically, let the KV cache be denoted as $\mathbf{K}, \mathbf{V} \in \mathbb{R}^{H \times L \times d}$, where $H$ is the number of heads, $L$ is the length of cache, and $d$ is the hidden dimension for each head. Let $C$ be the predefined chunk size, and define $\mathbf{K}_{[i]} := \mathbf{K}_{iC+1:(i+1)C+1} \in \mathbb{R}^{H \times C \times d}$ as the key cache corresponding to the $i$-th chunk (with similar notation for $\mathbf{V}_{[i]}$). 
Suppose the learnable query is denoted as $\mathbf{Q} \in \mathbb{R}^{H \times r \times d}$, where $r$ is a hyperparameter determining how many queries are used to summarize the KV cache in the current chunk. 
For each chunk, \modelname~performs multi-head cross-attention between $\mathbf{Q}$ and $\mathbf{K}_{[i]}, \mathbf{V}_{[i]}$ to obtain the summarized result $\mathbf{S}_{i}$ for chunk $i$:
\begin{equation}\label{eq:cross_attention}
    \mathbf{S}_{i}=\text{Softmax}\left(
    \frac{\mathbf{Q}\mathbf{K}_{[i]}}{\sqrt{d}}
    \right)\mathbf{V}_{[i]} \in \mathbb{R}^{r \times d_{kv}},
\end{equation}
where $d_{kv} = H \times d$ is the hidden state dimension after concatenation across all heads, and $i \in \left\{0, 1, \dots, L/C-1\right\}$.

After obtaining the chunk-wise results $\left\{\mathbf{S}_{0},\mathbf{S}_{1},\dots,\mathbf{S}_{L/C-1}\right\}$,
it is necessary to further aggregate the information across different chunks. 
This aggregation should consider locality, as the most recent information is likely to be more relevant to subsequent generation processes. Therefore, we employ a gated inter-chunk recurrence to aggregate this information:
 \begin{equation}\label{eq:inter_chunk_recurrent}
     \mathbf{h}_{i} = \mathbf{\alpha}_i\mathbf{h}_{i-1} + \mathbf{S}_{i} \in \mathbb{R}^{r\times d_{kv}},
 \end{equation}
where $\mathbf{h}_0$ is initialized to zero and $\alpha_i$ is a per-query scalar forget gate. 
Given recent research suggesting that data-dependent gating demonstrates more expressive power~\citep{beck2024xlstmextendedlongshortterm}, \modelname~determines each gating factor $\alpha_i$ with the following parameterization:
 \begin{equation}\label{eq:gating_factor}
     \mathbf{\alpha}_i= \sigma(\mathbf{S}_i\mathbf{W}_{\alpha}+b_{\alpha})^{\frac{1}{\tau}} \in \mathbb{R},
 \end{equation}
 where $\mathbf{W}_{\alpha}\in \mathbb{R}^{d_{kv} \times 1}$, $\sigma(\cdot)$ is sigmoid function, and $\tau=16$ is a temperature term that encourages the model to have a slower forgetting rate~\citep{Yang2023GatedLA}.
Through the data-dependent approach, the final context state $\mathbf{h}_{L/C-1}$ condenses the information from the KV cache. This condensed state, $\mathbf{h}_{L/C-1}$, is then integrated into the parameters of the pre-trained model using the low-rank adaptation method, thereby serving as the updated weight-level associative memory~\citep{Ramsauer2020HopfieldNI}.

\subsection{Weight Absorption}\label{sec:weight_absorption}
We expect the context states $\mathbf{h}$ to serve as newly learned knowledge from the context, which can be absorbed into the model’s weights. 
Drawing inspiration from the low-rank adaptation method~\cite{hu2021lora}, \modelname~assigns learnable queries to each linear layer in the pre-trained model and maps the KV cache corresponding to the block where the current linear layer resides to the context state using these queries. 
The parameters of the linear layer are then updated by integrating the context state with two low-rank matrices in a sandwich-like structure (Figure~\ref{fig:architecture}).

Specifically, a typical transformer-based LLMs is built by stacking a series of identical blocks, each containing a multi-head self-attention (MHA) layer and a FFN layer. 
Each block stores the KV cache computed by its MHA layer.
Therefore, for each parameter $\mathbf{W} \in \mathbb{R}^{d_{i}\times d_{o}}$ (where $d_i$ and $d_o$ denote the input and output dimensions, respectively) of the linear layer in the $l$-th block, and the stored KV cache $\mathbf{K}^{l}, \mathbf{V}^{l}$ of that block, \modelname~assigns a unique learnable query $\mathbf{Q}$ to each $\mathbf{W}$ and summarize $\mathbf{K}^{l}$ and $\mathbf{V}^{l}$ into the context state $\mathbf{h}$, following Equations~\ref{eq:cross_attention} and \ref{eq:inter_chunk_recurrent}.
This strategy of summarizing context with a unique query for each parameter allows the compression process to be adaptively learned from data for different weights.
Finally, two low-rank learnable matrices, $\mathbf{W}_{1} \in \mathbb{R}^{d_{i} \times d_{kv}}$ and $\mathbf{W}_{2} \in \mathbb{R}^{r \times d_{o}}$, are connected through $\mathbf{h}$ to absorb the context information into $\mathbf{W}$:
\begin{equation}\label{eq: absorb}
    \mathbf{W}^{'}=\mathbf{W}+\mathbf{W}_1\mathbf{h^T}\mathbf{W}_2.
\end{equation}
The second term, $\mathbf{W}_1\mathbf{h^T}\mathbf{W}_2$, can be interpreted as a simplified form of linear attention~\citep{schlag2021lineartransformerssecretlyfast} with context-informed keys and a fixed value prototype~\citep{caron2021unsupervisedlearningvisualfeatures}.
From this interpretation, the input $x \in \mathbb{R}^{d_i}$ is first projected to the KV dimension $d_{kv}$ via $\mathbf{W}_1$ and then used to compute the dot product similarity $x\mathbf{W}_1\mathbf{h}^T$ with the context state $\mathbf{h}$. 
This similarity weights the pre-learned prototype $\mathbf{W}_2$ and produces the output that is updated through the new context-informed weight.

In implementation, considering that the KV dimension $d_{kv}$ is typically large in current LLMs (e.g., 1024 for LLaMA-3-8B), resulting in a significant number of new learnable parameters, \modelname~introduces an additional down-projection learnable parameter $\mathbf{W}_{down} \in \mathbb{R}^{H \times d \times d'}$ (where $d' \ll d$) to reduce the parameter size by down-projecting the $\mathbf{V}$ cache:

\begin{equation}\label{eq:v_proj_low}
\mathbf{V}' = \mathbf{V} \mathbf{W}_{down} \in \mathbb{R}^{H \times L \times d'}.
\end{equation}

As a result, the original dimension $d_{kv} = H \times d$ is projected to $d_{kv}' = H \times d'$, thereby reducing the number of the learnable parameters.

\begin{figure}[t!]
  \centering
   \includegraphics[width=0.9\textwidth]{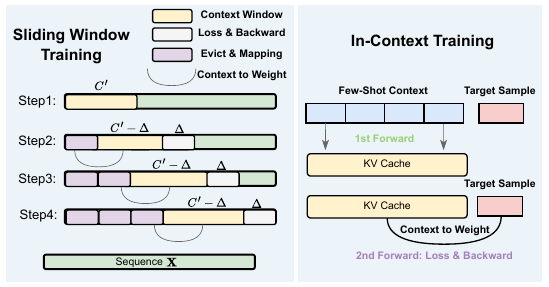}  
    \caption{Training strategy of \modelname. The sliding-window strategy accumulates loss from each step in a sequence and updates \modelname's parameters after the entire sequence has been processed. The in-context training employs a 2-forward-1-backward strategy: the first forward pass computes the KV cache without gradient computation, while the second forward pass updates the model parameters using the KV cache from the first forward pass and calculates the loss to update the parameters introduced by \modelname}
    \vspace{-5mm}
  \label{fig:training_strategy}
\end{figure}

\subsection{Training Strategy}
\modelname's reliance on the KV cache for parameter updates necessitates a departure from conventional next-token prediction training methods. 
To address this, we have developed two distinct training strategies tailored to the specific requirements of language generation and language understanding tasks: sliding window training and in-context training (Figure~\ref{fig:training_strategy}).

\subsubsection{Sliding Window Training}\label{sec:sliding_window_training}
For general language generation tasks, we employ a sliding window strategy to train \modelname~for mapping context into parameter updates. This process begins with general language corpora, which we divide into sequences $\mathbf{X}$ of length $L$. 
For each sequence, we utilize a window size $C'$ and a stride size $\Delta$. It's important to note that $C'$ here is larger than the context length $C$ used in Section~\ref{sec:context_mapping}.
We start by initializing the window with the first $C'$ tokens of $\mathbf{X}$.
Then, we begin an iterative process. In each step, we evict the earliest $\Delta$ tokens. 
The KV caches of these evicted tokens are then used to generate parameter updates. 
Simultaneously, we calculate the next token prediction loss for the incoming $\Delta$ tokens using the updated parameters.
As we progress through the sequence, the loss is accumulated until the entire sequence $\mathbf{X}$ has been processed. 
Once the sequence is fully seen, we update \modelname's parameters using this accumulated loss.
This sliding window approach enables \modelname~to efficiently process long sequences while maintaining a fixed context size. By continuously updating the window and accumulating loss, the model learns to utilize context information effectively across various positions in the input sequence.

\subsubsection{In-Context Training}\label{sec:multi_task_training}
To adapt \modelname~for language understanding tasks, we employ an in-context training strategy using a selected set of tasks.
For each sample in each task's training set, we first randomly sample $k$ examples to form a few-shot context and store their KV caches (1st forward pass without gradient computation). 
We then update the base model parameters using this cache and compute the loss for the current sample to optimize the parameters introduced by \modelname~(2nd forward pass with backpropagation).

\subsection{Inference Strategy}
For model inference, inspired by context-locality \citep{li2024snapkvllmknowslooking,zhang2023h2oheavyhitteroracleefficient}, we adopt a hybrid approach tailored to different task types:
For language understanding tasks, we convert most demonstrations into weight updates, retaining only a small portion of recent context.
For long context generation tasks, we use a sliding window strategy with stride size $\Delta$ smaller than window size $C'$. We keep the most recent context intact while transforming the evicted context into temporary model updates via \modelname.
This adaptive strategy balances immediate context and adapted knowledge from earlier inputs, optimizing efficiency and performance across different scenarios.
\section{Experiments and Results}
We evaluate \modelname~across various model scales and architectures, focusing on both language understanding tasks and language generation tasks. 
We also explore the scaling ability of \modelname~with different numbers of in-context demonstrations across various tasks and lengths. 
Additionally, We evaluate \modelname's efficiency and robustness through comprehensive ablation studies and in-depth analyses.
\subsection{Experimental Setting}
\paragraph{Base Model}
we select TinyLlama-1.1B~\citep{zhang2024tinyllamaopensourcesmalllanguage}, LLaMA-3-8B, and Phi-3-Medium~\citep{abdin2024phi3technicalreporthighly} as our base model.
In all experiments, we froze the original model weights and only trained the parameters introduced by \modelname.

\paragraph{Base Setting}
Without explicit specialization, we apply \modelname~to every linear layer of the pre-trained model. 
The default chunk size $C$ in Section~\ref{sec:context_mapping} is set to 128, and the down-projected value dimension $d_{kv}'$ in Equation~\ref{eq:v_proj_low} is set to 32 for all base models. 
When performing chunk-wise cross-attention, in cases where the input KV cache is not divisible by $C$, we performe an additional cross-attention operation on the remaining KV cache after division and concatenate the result with the chunk-wise result. 
The number of learnable queries in \modelname~is set to 16 for TinyLlama-1.1B and LLaMA-3-8B, and 48 for Phi-3-Medium.

\subsection{Language Understanding Task}\label{sec:language_under_task}
\paragraph{Training Details}
For adapting \modelname~to language understanding tasks, we employ the in-context training approach introduced in Section~\ref{sec:multi_task_training}. We carefully select several tasks for training.
The tasks included BoolQ \citep{clark2019boolq}, CoPA \citep{roemmele2011choice}, SST2 \citep{socher-etal-2013-recursive}, CB \citep{demarneffe:cb}, and RTE \citep{bentivogli2009fifth}.
For each sample in training set across all tasks, we randomly select context examples from the training set for computing the KV cache in the first forward pass. The number of demonstrations tailored to each model's capacity: 30 samples for TinyLlama-1.1B, and 60 samples for both LLaMA-3-8B and Phi-3-Medium.
For further training detailes, please refer to Appendix~\ref{sec:train_lu}.

\paragraph{Evaluation and Baseline}
We evaluate \modelname~across a diverse set of language understanding tasks, including both those encountered during the training stage and unseen tasks. 
Our comparison encompasses several baseline methods: zero-shot prompting, ICL, and two heuristic context eviction strategies—SnapKV \citep{li2024snapkvllmknowslooking} and $\text{H}_2\text{O}$ \citep{zhang2023h2oheavyhitteroracleefficient}. We also include TempLoRA \citep{wang2024greatertextcomesgreater}, a test-time low-rank adaptation method, as a baseline. For a fair comparison, we additionally incorporate results obtained after fine-tuning the base model using LoRA \citep{hu2021lora} with the same number of learnable parameters as \modelname.
For detailed settings of the different methods, please refer to Appendix~\ref{sec:eval_lu}.

\begin{table}[t!]
\centering
\caption{Evaluation results on language understanding tasks after in-context training. OBQA: OpenbookQA. ARC-C: ARC-Challenge. ARC-E: ARC-Easy}
\label{tab:multitask}
\resizebox{\columnwidth}{!}{%
\begin{tabular}{l|c|cccccc|ccccccc}
\hline
                       &                                  & \multicolumn{6}{c|}{Seen Task}                                                                      & \multicolumn{7}{c}{Unseen Task}                                                                                      \\ \hline
                       &                                  & BoolQ          & CoPA           & SST2           & CB             & RTE            & Avg.           & Hellaswag      & Winogrande     & OBQA           & ARC-C          & ARC-E          & PIQA           & Avg.           \\ \hline
Zero-shot              & \multirow{7}{*}{Tiny-Llama-1.1B} & 57.03          & \textbf{78.00} & 69.61          & 14.29          & 51.99          & 54.18          & 59.01          & 58.88          & 21.80          & 27.56          & 60.31          & 73.34          & 50.15          \\
ICL$_{10-\text{shot}}$ &                                  & 63.39          & 76.00          & 78.21          & 51.79          & 49.10          & 63.70          & \textbf{59.46} & 59.83          & 26.20          & 30.61          & 64.81          & \textbf{73.78} & 52.45          \\
LoRA                   &                                  & 74.28          & \textbf{78.00} & 86.58          & 80.36          & 69.31          & 77.71          & 59.01          & 56.35          & 24.80          & 27.90          & 57.15          & 72.14          & 49.56          \\
TempLoRA               &                                  & 57.61          & 74.00          & 71.33          & 39.29          & 56.68          & 59.78          & 59.42          & 60.46          & 21.60          & 27.82          & 62.12          & 72.09          & 50.59          \\
H$_2$O                 &                                  & 62.72          & 77.00          & 75.80          & 23.21          & 47.23          & 57.19          & 58.86          & 59.12          & 23.20          & 30.20          & 62.58          & 72.96          & 51.15          \\
SnapKV                 &                                  & 63.29          & 75.00          & 63.30          & 44.64          & 46.93          & 58.63          & 58.09          & 59.83          & 23.40          & 28.75          & 62.92          & 72.80          & 50.97          \\ \rowcolor{cyan!50}
StreamAdapter          &                                  & \textbf{81.77} & 76.00          & \textbf{89.56} & \textbf{85.71} & \textbf{82.67} & \textbf{83.14} & 59.45          & \textbf{59.91} & \textbf{27.00} & \textbf{31.06} & \textbf{64.93} & 73.29          & \textbf{52.61} \\ \hline
Zero-shot              & \multirow{7}{*}{LLaMA-3-8B}      & 81.11          & 88.00          & 67.09          & 51.79          & 67.87          & 71.17          & 79.15          & 72.93          & 35.60          & 50.09          & 80.35          & 79.60          & 66.29          \\
ICL$_{10-\text{shot}}$ &                                  & 83.49          & 89.00          & 95.07          & 83.93          & 79.06          & 86.11          & \textbf{81.46} & 78.45          & 36.40          & 52.43          & 82.79          & 80.58          & 68.69          \\
LoRA                   &                                  & 89.24          & 84.00          & 96.22          & 96.43          & 88.09          & 90.80          & 80.78          & 69.14          & 35.40          & 51.62          & 80.77          & 78.67          & 66.06          \\
TempLoRA               &                                  & 81.77          & 91.00          & 90.60          & 76.79          & 70.40          & 82.11          & 79.62          & 76.56          & 34.20          & 50.85          & 81.06          & 79.27          & 66.93          \\
H$_2$O                 &                                  & 82.48          & 88.00          & 92.43          & 73.21          & 75.81          & 82.39          & 80.57          & 75.77          & 33.80          & 51.45          & 82.62          & 79.98          & 67.37          \\
SnapKV                 &                                  & 84.46          & 88.00          & 94.41          & 75.00          & 74.01          & 83.18          & 80.41          & 76.95          & 35.20          & 50.77          & 82.74          & 79.65          & 67.62          \\ \rowcolor{cyan!50}
StreamAdapter          &                                  & \textbf{90.15} & \textbf{89.00} & \textbf{96.72} & \textbf{98.21} & \textbf{89.67} & \textbf{92.75} & 80.87          & \textbf{78.64} & \textbf{38.80} & \textbf{53.03} & \textbf{83.53} & \textbf{80.59} & \textbf{69.24} \\ \hline
Zero-shot              & \multirow{7}{*}{Phi-3-Medium}    & 88.65          & 92.00          & 90.48          & 73.21          & 80.51          & 84.97          & 79.12          & 76.48          & 37.80          & 54.95          & 80.85          & 80.90          & 68.35          \\
ICL$_{10-\text{shot}}$ &                                  & 88.23          & \textbf{94.00} & 95.07          & 83.93          & 77.62          & 87.77          & 80.70          & \textbf{79.40} & \textbf{44.80} & 62.80          & 86.87          & \textbf{82.37} & 72.82          \\
LoRA                   &                                  & 88.59          & 92.00          & \textbf{96.22} & \textbf{92.86} & 89.53          & 91.84          & 79.36          & 76.87          & 41.80          & 58.60          & 84.22          & 81.18          & 70.34          \\
TempLoRA               &                                  & 89.27          & 92.00          & 93.81          & 71.43          & 78.34          & 84.97          & 79.30          & 76.56          & 44.40          & 60.58          & 86.28          & 80.61          & 71.29          \\
H$_2$O                 &                                  & 83.46          & \textbf{94.00} & 94.95          & 80.46          & 79.78          & 86.53          & 80.61          & 76.11          & 40.80          & 62.12          & 86.41          & 82.07          & 71.35          \\
SnapKV                 &                                  & 78.99          & 89.00          & 70.41          & 80.36          & 68.95          & 77.54          & 80.21          & 72.53          & 41.30          & 61.71          & 83.89          & 79.87          & 69.92          \\ \rowcolor{cyan!50}
StreamAdapter          &                                  & \textbf{90.24} & 92.00          & 95.18          & \textbf{92.86} & \textbf{89.89} & \textbf{92.03} & \textbf{82.03} & 77.11          & \textbf{44.80} & \textbf{64.33} & \textbf{86.91} & 82.47          & \textbf{72.94} \\ \hline
\end{tabular}%
}
\end{table}

\begin{figure}[t!]
  \centering
   \includegraphics[width=\textwidth]{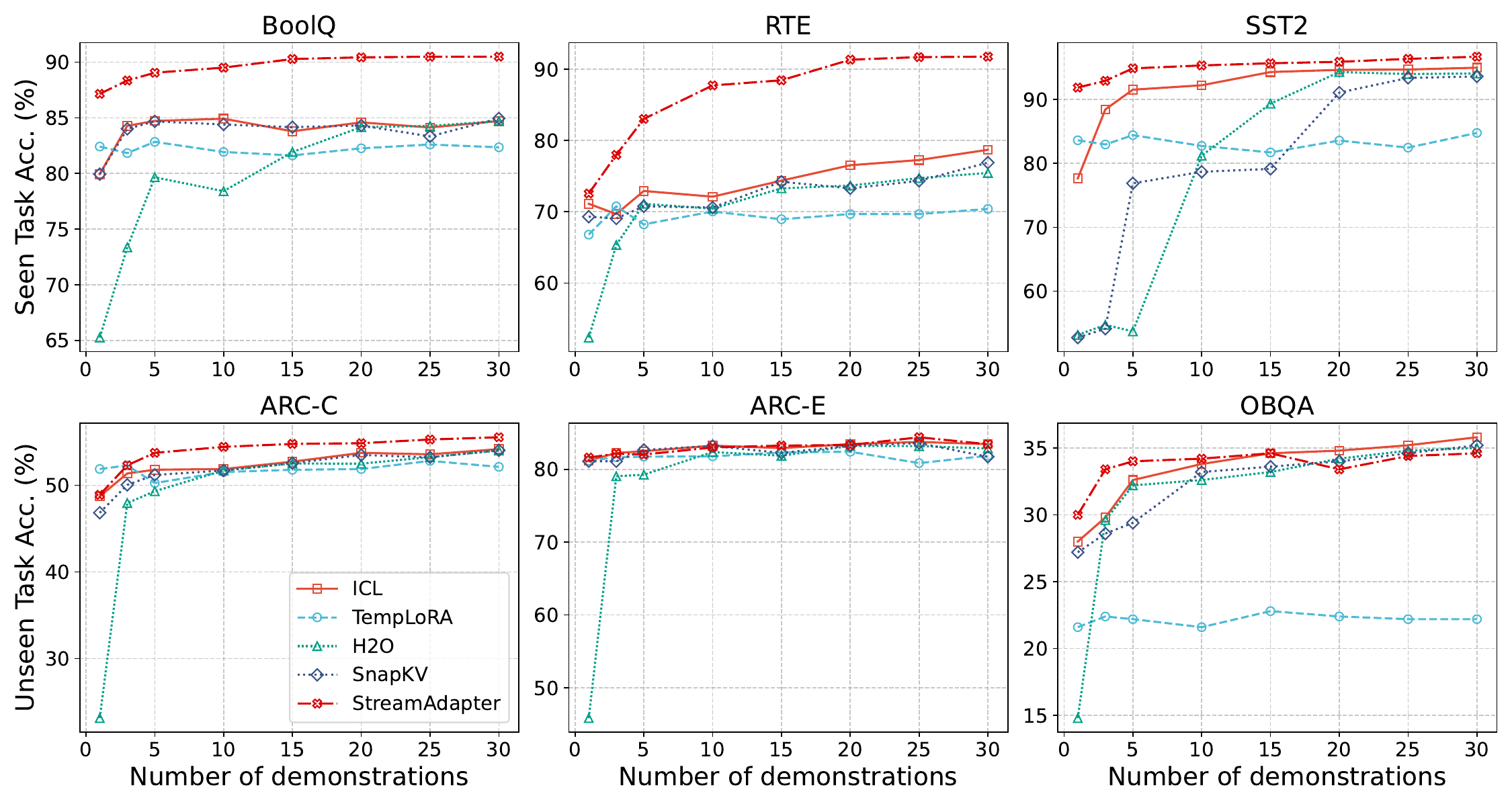}  
    \caption{Comparison of various methods across different tasks with different numbers of demonstrations}
  \label{fig:de_vs_pe}
\end{figure}

\paragraph{Evaluation Result}
The evaluation results in Table~\ref{tab:multitask} show that \modelname~consistently outperforms LoRA on the test set across the seen tasks, despite using the same training recipe and parameter count. 
On unseen tasks, \modelname~also surpasses all other methods across the three models. Unlike context selection methods such as SnapKV and H$_2$O, which are upper-bounded by full ICL, \modelname~enhances model capability by absorbing context and even outperforms full ICL. Additionally, while LoRA exhibits performance degradation on unseen tasks, indicating catastrophic forgetting, \modelname~maintains improved results, demonstrating its effectiveness and generalization capabilities. 

\paragraph{Scaling Analysis}
We examine the adaptation accuracy of various methods, including \modelname, as the number of demonstrations increases across different tasks using the LLaMA-3-8B model.
To ensure fair comparison across methods, we employ a consistent approach to demonstration selection. 
For each task, we generate a fixed set of demonstrations from its training set. All test samples are then evaluated using this same set of demonstrations across all methods.
For detailed configuration information, please refer to Appendix~\ref{appendix:dvp_setting}.

Figure~\ref{fig:de_vs_pe} shows that \modelname~consistently improves with more demonstrations on both seen and unseen tasks. 
On both seen and unseen tasks, \modelname~significantly outperforms TempLoRA and achieves better results than ICL and other context eviction strategies. 
The increasing accuracy with more demonstrations, particularly on unseen tasks, suggests that \modelname~effectively leverages contextual information to encode knowledge into parameters rather than simply memorizing task-specific patterns.

These results highlight \modelname's potential as a robust approach for TTA in language models, demonstrating its ability to generalize across diverse language understanding tasks.

\subsubsection{Language Generation Task}\label{sec:language_generation}
\paragraph{Training Details}
For training \modelname~on language generation tasks, we utilize the training set of the PG19 dataset~\citep{raecompressive2019}, employing the sliding window strategy introduced in Section~\ref{sec:sliding_window_training}.
The sequence length $L$ is set to 8192 for TinyLlama-1.1B, and 16384 for LLaMA-3-8B and Phi-3-Medium. 
The SW size $C'$ for all models is fixed at 1024, with a stride $\Delta$ of 512. 
For additional training hyperparameters, please refer to Appendix~\ref{sec:train_lg}.

\begin{table}[t!]
\centering
\caption{Comparison on PG19 test set across varying maximum context lengths using sliding window evaluation strategy}
\label{tab:pg19_split}
\resizebox{0.8\columnwidth}{!}{%
\begin{tabular}{l|c|ccccc}
\hline
                                &                & 16K   & 32K   & 64K   & 128K  & 256K  \\ \hline
\multirow{3}{*}{TinyLlama-1.1B} & Sliding Window & 11.16 & 11.24 & 11.29 & 11.29 & 11.28 \\ 
                                & TempLoRA       & 11.04 & 11.03 & 10.99 & 10.93 & 10.91 \\ \rowcolor{cyan!50}
                                & StreamAdapter  & \textbf{10.81} & \textbf{10.88} & \textbf{10.70} & \textbf{10.44} & \textbf{9.97}  \\ \hline
\multirow{3}{*}{LLaMA-3-8B}     & Sliding Window & 9.44  & 9.52  & 9.51  & 9.50  & 9.50  \\
                                & TempLoRA       & 9.73  & 9.81  & 9.81  & 9.83  & 9.84  \\ \rowcolor{cyan!50}
                                & StreamAdapter  & \textbf{9.27}  & \textbf{9.33}  & \textbf{9.23 } & \textbf{9.22}  & \textbf{9.21}  \\ \hline
\end{tabular}%
}
\end{table}
\begin{figure}[t!]
  \centering
   \includegraphics[width=\textwidth]{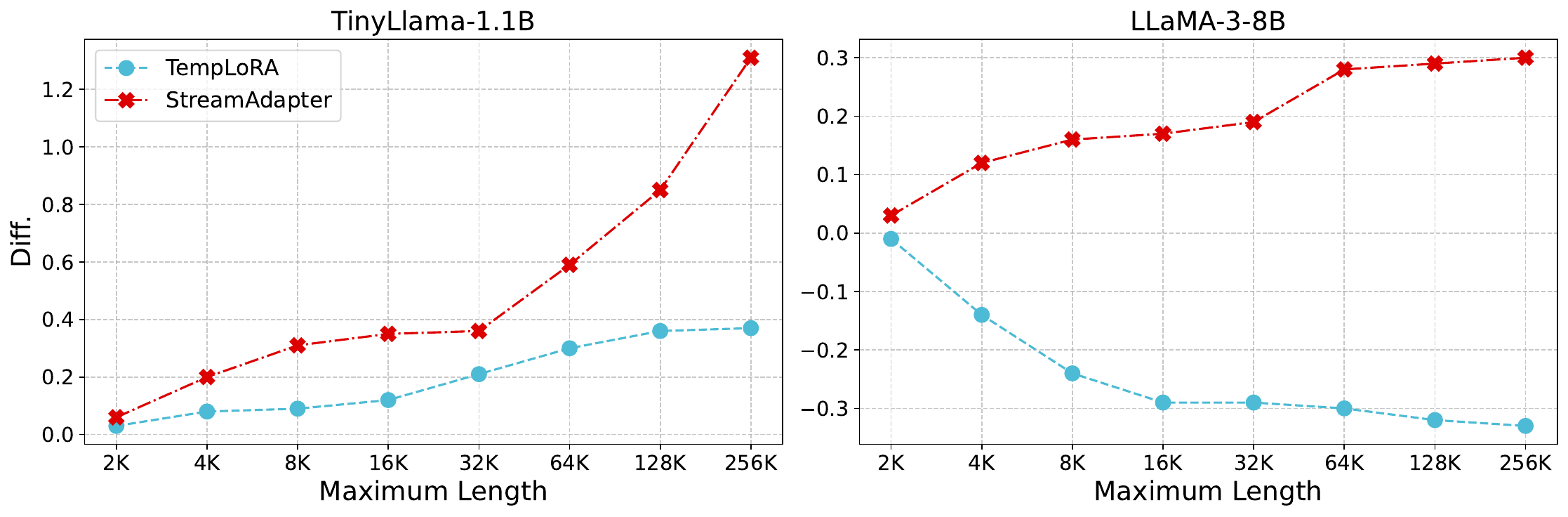}  
    \caption{Perplexity gap between TTA methods and sliding window strategy across varying maximum context lengths on the PG19 test set}
  \label{fig:pg19_diff}
\end{figure}

\paragraph{Evaluation and Baselines}
We evaluate \modelname~on the PG19 test set using various maximum truncation lengths. For each sample, we employ the sliding window evaluation strategy with a window size $C'$ of 1024 and a stride $\Delta$ of 512. Perplexity is computed in the incoming stride window, and we report the average perplexity across the entire test set. For comparison, we use two baselines: naive sliding window approach with identical settings, and TempLoRA. 
Detailed parameter settings for TempLoRA are provided in Appendix~\ref{sec:eval_lg}.

\paragraph{Evaluation Result}
The results are presented in Table~\ref{tab:pg19_split}. They clearly demonstrate that \modelname~outperforms both the sliding window approach and TempLoRA across all maximum context length.
Notably, while TempLoRA achieves lower generation perplexity than sliding window with TinyLlama-1.1B, it shows inferior performance when evaluated with LLaMA-3-8B. We hypothesize that this discrepancy may be due to LLaMA-3-8B's training on high-quality corpora. 
TTA with TempLoRA on the current chunk might lead LLaMA-3 to overfit to that chunk, resulting in inaccurate predictions on subsequent context.
In contrast, \modelname~exhibits superior generation performance on both TinyLlama-1.1B and LLaMA-3-8B models, showcasting its wide applicability across different model scales. 
Moreover, we visualize the perplexity gap between \modelname~and TempLoRA compared to the sliding window approach at different maximum lengths in Figure~\ref{fig:pg19_diff}. 
The gap consistently widens as the context size increases for both TinyLlama-1.1B and LLaMA-3-8B models.
The consistent improvement across increasing context lengths highlights \modelname's ability to effectively leverage additional contextual information, regardless of the base model's scale. This scalability further emphasizes \modelname's robustness and adaptability in processing long-form text, making it particularly suitable for applications requiring efficient handling of extensive contextual data.

\subsection{Analysis}\label{sec:analysis}
\paragraph{Efficiency}
\begin{figure}[t!]
\centering
\includegraphics[width=\textwidth]{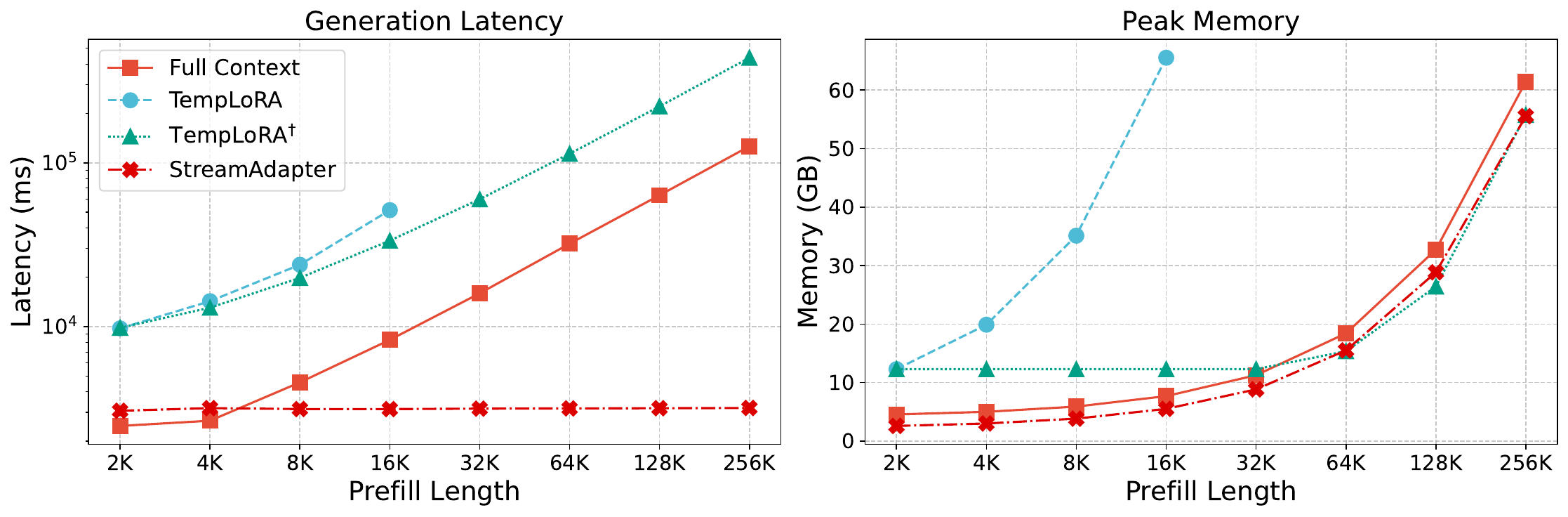}
\caption{Generation latency and peak memory consumption across different prefill lengths. $\dagger$ indicates adaptation using sequential chunk-wise strategy, as directly mapping all prefill context leads to out-of-memory}
\label{fig:efficiency}
\end{figure}

We compare the end-to-end latency and peak memory consumption of model generation with TinyLlama-1.1B across various prefill context lengths. 
Our evaluation process begins by generating the KV cache for a given prefill context length, followed by measuring the latency of generating 128 tokens using three methods: full context, TempLoRA, and our \modelname. 

The hyperparameter settings for TempLoRA and \modelname~are consistent with those described in Section~\ref{sec:language_generation}. 
All results are averaged across five runs with a single NVIDIA A100-80G GPU.

The results, presented in Figure~\ref{fig:efficiency}, clearly demonstrate that \modelname~maintains constant generation latency across different prefill context lengths (i.e., different KV cache sizes). 
In contrast, the latency of full context generation and TTA with TempLoRA increases almost linearly with the context size.
Moreover, TempLoRA's need for gradient backpropagation during adaptation leads to substantial GPU memory consumption as the prefill context increases. While this can be mitigated using sequential chunk-wise adaptation (with a chunk size of 2048 in our setting), this approach increases the generation latency.
Conversely, \modelname's recurrent design allows simultaneous mapping of all context without requiring sequential chunk-wise processing. 
Although \modelname's peak memory consumption also increases with larger prefill contexts, we attribute this to the current implementation materializing all intermediate states. As only the final state is needed, we believe further optimizations, similar to those in \citep{gu2023mamba}, could reduce \modelname's memory demands.

\paragraph{Adaptation Ratio}
\begin{figure}[t!]
\centering
\includegraphics[width=\textwidth]{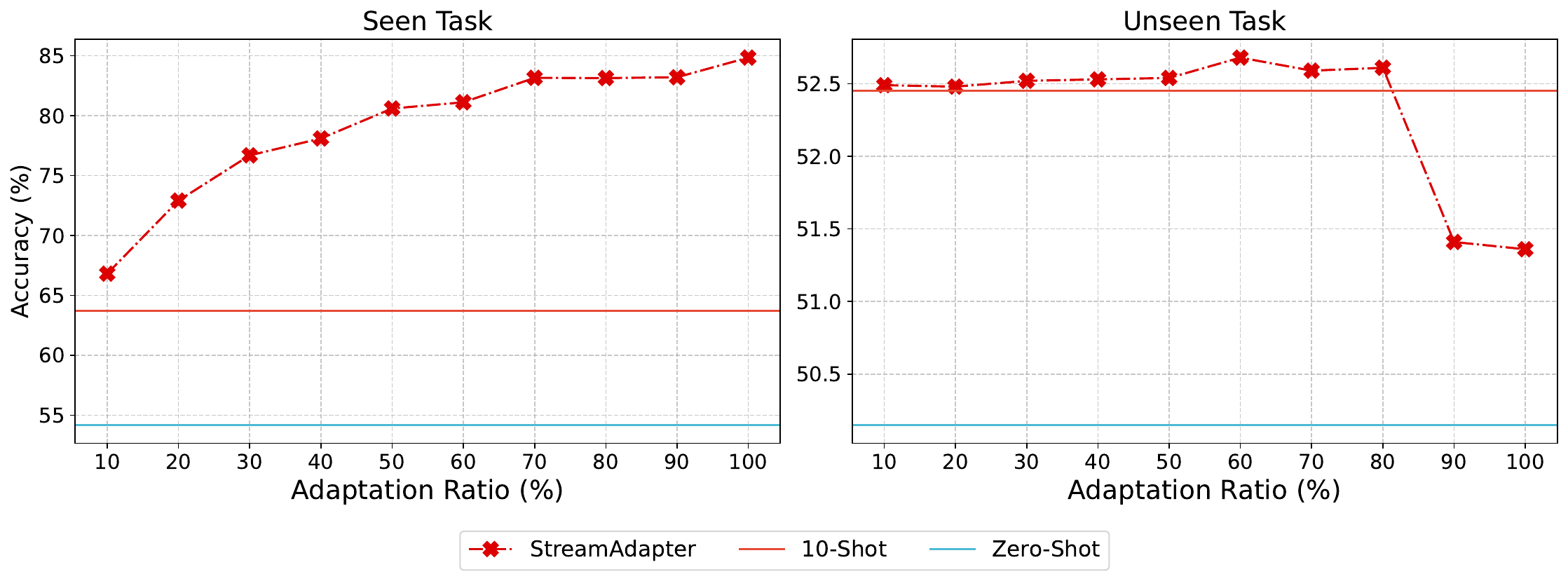}
\caption{Average accuracy of \modelname~with TinyLlama-1.1B across different adaptation ratios on both seen and unseen tasks}
\label{fig:ana_ratio}
\end{figure}
In the language understanding tasks described in Section~\ref{sec:language_under_task}, we adapt a fixed ratio of context into model weights and evaluate the model with the remaining context in context. To explore the relationship between adaptation ratio and final accuracy on both seen and unseen tasks, we evaluate TinyLlama-1.1B with fixed 10-shot samples across different adaptation ratios.

The results of our adaptation ratio analysis are presented in Figure~\ref{fig:ana_ratio}. 
For a more detailed breakdown of acuracy on each individual task, please refer to Appendix~\ref{appendix:detail_adaptation}.
\modelname~generally performs better on seen tasks when adapting more demonstrations. 
For unseen tasks, \modelname~outperforms 10-shot ICL when adapting 10\%-80\% of demonstrations but shows a decline with extreme adaptation ratios (90\% or 100\%). 
Although teh adaptation accuracy remains better than zero-shot prompting, we hypothesize that retaining a small portion of demonstrations is necessary to guide adaptation direction on unseen tasks. 
This is likely because \modelname~learns mapping relations from a limited set of tasks and may adapt the base model in a direction different from the target unseen task. We posit that training \modelname~with a more diverse task set could address this issue, which we leave for future work.

\paragraph{Robustness}
\begin{table}[t!]
\centering
\caption{Evaluation results on language understanding tasks with different prompt templates for in-context examples and evaluated samples}
\label{tab:lu_analysis}
\resizebox{0.9\columnwidth}{!}{%
\begin{tabular}{l|cccl|cccl}
\hline
                & \multicolumn{4}{c|}{Seen Task}                                    & \multicolumn{4}{c}{Unseen Task}                                   \\ \hline
                & BoolQ          & SST2           & RTE            & Avg.           & ARC-C          & ARC-E          & PIQA           & Avg.           \\ \hline
ICL$_{10-shot}$ & 53.06          & 50.92          & 48.10          & 50.69          & 25.17          & 50.13          & 72.80          & 49.37          \\ \rowcolor{cyan!50}
StreamAdapter   & \textbf{71.38} & \textbf{50.92} & \textbf{82.03} & \textbf{68.11} & \textbf{25.43} & \textbf{50.67} & \textbf{72.96} & \textbf{49.69} \\ \hline
\end{tabular}%
}
\end{table}

We evaluate the influence of using different templates for in-context examples and target evaluated samples to analyze the robustness of different TTA methods as patterns change. For this analysis, we use the TinyLlama-1.1B model trained from Section~\ref{sec:language_under_task}.
We select three seen tasks (BoolQ, SST2, RTE) and three unseen tasks (ARC-Challenge \citep{clark2018thinksolvedquestionanswering}, ARC-Easy \citep{clark2018thinksolvedquestionanswering}, PIQA \citep{Bisk2020}) to verify the robustness of full in-context learning (ICL) and \modelname. We fix the number of in-context examples at 10, with other details for \modelname~remaining the same as in Section~\ref{sec:language_under_task}.

\begin{figure}[t!]
\centering
\includegraphics[width=\textwidth]{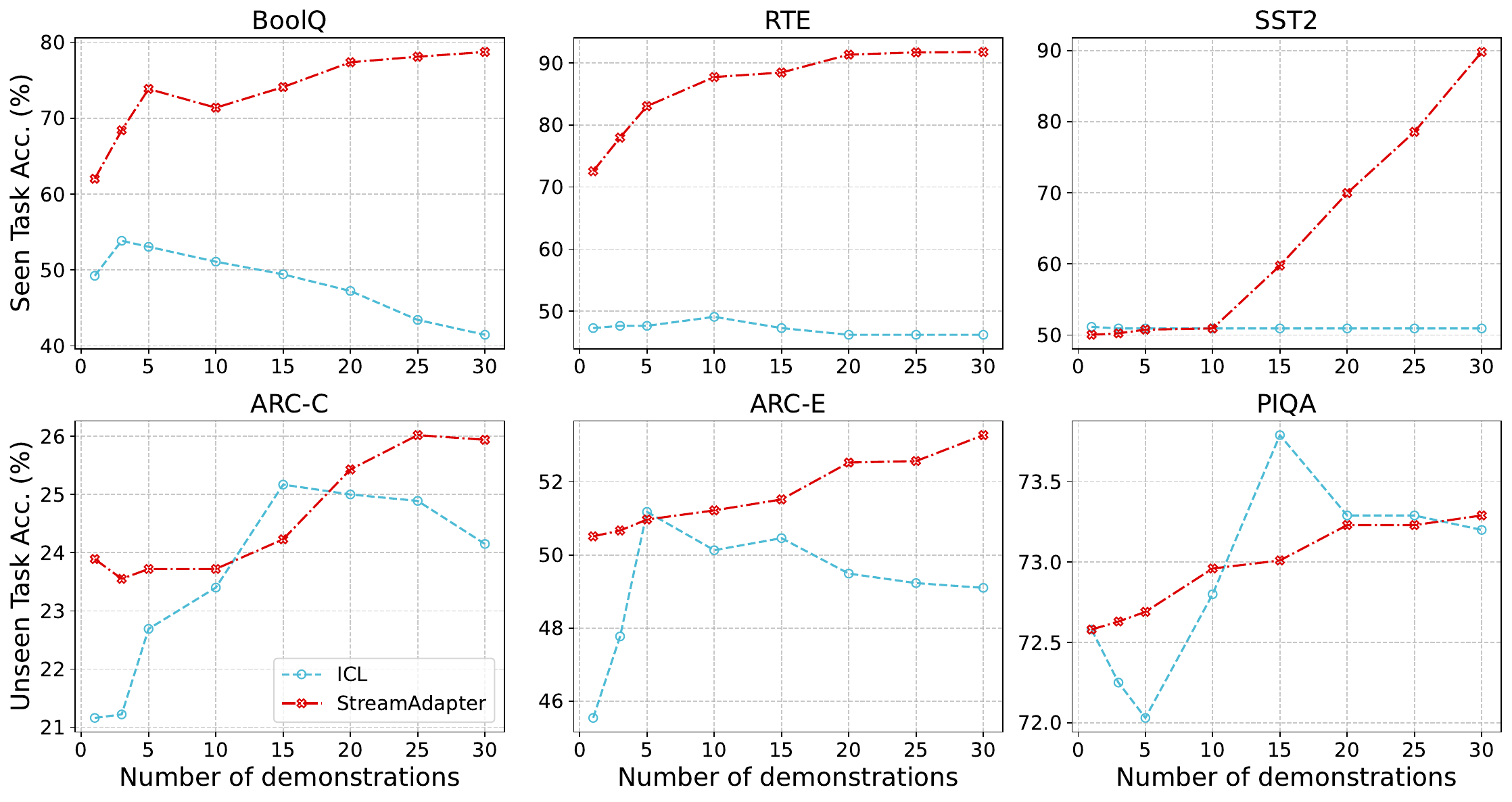}
\caption{Evaluation with varying numbers of demonstrations, using different prompt templates for evaluated samples and in-context examples}
\label{fig:analysis}
\end{figure}

The results, presented in Table~\ref{tab:lu_analysis}, show that although both full ICL and \modelname~exhibit degraded accuracy compared to Table~\ref{tab:multitask}, \modelname~still outperforms ICL on both seen and unseen tasks.
Moreover, as illustrated in Figure~\ref{fig:analysis}, ICL's average accuracy decreases as the number of in-context examples increases, suggesting that ICL primarily memorizes patterns and fails to adapt when these patterns change.
Conversely, \modelname~consistently achieves higher accuracy with additional demonstrations, indicating that TTA with \modelname~leverages contextual information to enhance model capability rather than simply memorizing task-specific patterns.

\begin{table}[t!]
    \centering
    \begin{minipage}{.35\textwidth}
        \centering
        \caption{Evaluation on language understanding task with different chunk size}
        \label{tab:ab_chunk}
        \vspace{1mm}
        \resizebox{\textwidth}{!}{%
        \begin{tabular}{@{}lcc@{}}
        \toprule
        ChunkWise & Seen Task Avg. & Unseen Task Avg. \\ \midrule
        No Chunking    & 82.84          & 51.46            \\
        64        & 82.23          & 51.47            \\
        128       & \textbf{86.99} & \textbf{51.92}   \\
        256       & 86.10          & 51.71            \\ \bottomrule
        \end{tabular}%
        }
    \end{minipage}%
    \hfill
    \begin{minipage}{.53\textwidth}
        \centering
        \caption{Evaluation on language understanding task with different number of learnable query for summarizing each chunk}
        \label{tab:ab_query}
        \vspace{1mm}
        \resizebox{\textwidth}{!}{%
        \begin{tabular}{@{}cccc@{}}
        \toprule
        \# of Query & \# of Learnable Param. (\%) & Seen Task Avg. & Unseen Task Avg. \\ \midrule
        16         & 4.99                      & 83.14          & \textbf{52.61}   \\
        32         & 5.55                      & 86.50          & 52.01            \\
        64         & 6.66                      & 87.21          & 51.99            \\
        128        & 8.81                      & \textbf{87.22} & 52.05            \\ \bottomrule
        \end{tabular}%
        }
    \end{minipage}
\end{table}

\begin{table}[t!]
\centering
\caption{Evaluation results on language understanding task with fixed chunk size / query ratio using TinyLlama-1.1B}
\label{tab:holistic_eval}
\resizebox{0.8\columnwidth}{!}{%
\begin{tabular}{ccccc}
\hline
Chunk Size & \# of Query & \# Learnable Param. & Seen Task Avg. & Unseen Task Avg. \\ \hline
64         & 8           & 4.20                & 85.33          & 52.30            \\
128        & 16          & 4.99                & 83.14          & 52.61            \\
256        & 32          & 5.55                & 87.80          & 52.50            \\ \hline
\end{tabular}%
}
\end{table}

\subsection{Ablation}
We examine the impact of different components and settings of \modelname, focusing our analysis on the TinyLlama-1.1B model and evaluating its adaptation capability on language understanding tasks. Except for the specific parameter settings under investigation, all other training and evaluation settings remain consistent with those described in Section~\ref{sec:language_under_task}.

We begin by examining the effectiveness of the chunk-wise design and the influence of chunk size on \modelname's performance. For this analysis, we fix the number of queries used to compress each chunk at 16. In the absence of a chunk-wise design, we would directly summarize the entire KV cache using queries with cross-attention to generate new model weights, eliminating the need for inter-chunk recurrence.
Table~\ref{tab:ab_chunk} shows that the chunk-wise approach outperforms the non-chunked version, with a chunk size of 128 achieving the best results on both seen (86.99\%) and unseen tasks (51.92\%). 
Smaller (64) and larger (256) chunk sizes show suboptimal results, indicating that 128 strikes the right balance in capturing contextual information with 16 queries.

Next, we examine the effect of different numbers of queries, with results presented in Table~\ref{tab:ab_query}. 
As the number of queries per chunk increases, accuracy improves on seen tasks but declines on unseen tasks. We hypothesize that increasing the number of learnable parameters through additional queries causes \modelname~to trend towards memorizing fixed patterns from training tasks, resulting in poorer generalization to unseen tasks.

From the results in Tables~\ref{tab:ab_chunk} and~\ref{tab:ab_query}, we hypothesize that the optimal ratio of context tokens per query is 128 / 16 = 8. To validate this hypothesis, we conduct experiments with different chunk sizes while maintaining this fixed ratio.
The results from Table~\ref{tab:holistic_eval} show that maintaining this ratio does improve the adaptation accuracy on both seen and unseen tasks compared to the results from Table~\ref{tab:ab_chunk}. 
However, the highest accuracy is still achieved with the original chunk size of 128 and 16 queries.
We hypothesize that this optimal configuration may be related to the context length of each task's in-context examples (presented in Table~\ref{tab:data_stat}). Further analysis of this relationship is left for future work.
\section{Conclusion}
We introduce \modelname, a novel approach for adapting pretrained LLMs at test time directly from given context. 
\modelname~employs context mapping and weight absorption mechanisms to efficiently transform context tokens into parameter updates, achieving similar or superior results to full-context generation while reducing both memory consumption and inference time.
Evaluations across diverse language understanding and generation tasks with various model scales demonstrate \modelname's effectiveness in adapting to new tasks, outperforming fine-tuning and zero-shot prompting, while also surpassing full ICL. 
Analysis reveals \modelname's superior scalability and robustness across varying context lengths and adaptation ratios, while maintaining constant inference time and memory consumption.
These promising results open new avenues for efficient TTA of LLMs, paving the way for more flexible and customized language model deployments. 
Future work could explore \modelname's application to more diverse tasks and larger model scales, potentially extending its principles to other modalities.

\bibliography{main}
\bibliographystyle{abbrvnat}
\appendix
\section{Training Details}
\subsection{Language Understanding Task}\label{sec:train_lu}
We use the training sets of BoolQ \citep{clark2019boolq}, CoPA \citep{roemmele2011choice}, SST2 \citep{socher-etal-2013-recursive}, CB \citep{demarneffe:cb}, and RTE \citep{bentivogli2009fifth} for training on language understanding tasks. 
We construct each sample with pre-defined template, the template for each task is presented in Table~\ref{tab:template_training}.

For training \modelname, we employ the WarmupCosine learning rate scheduler and the AdamW optimizer with $(\beta_1, \beta_2) = (0.9, 0.95)$ and weight decay 0.01 for 3 epochs. The hyperparameters vary across models: for TinyLlama-1.1B, we use a batch size of 16, learning rate of $5 \times 10^{-5}$, and 100 warmup steps; for LLaMA-3-8B, a batch size of 4, learning rate of $2 \times 10^{-5}$, and 500 warmup steps; and for Phi-3-Medium, a batch size of 2, learning rate of $1 \times 10^{-5}$, and 800 warmup steps.

For training LoRA, we apply the adapter to every linear layer of the pre-trained model and use the same learning rate scheduler, optimizer, and number of epochs as for training \modelname. 
The rank and $\alpha$ of LoRA are both set to 64.
The hyperparameters are adjusted for each model: TinyLlama-1.1B uses a batch size of 16, learning rate of $1 \times 10^{-4}$, and 100 warmup steps; LLaMA-3-8B uses a batch size of 8, learning rate of $8 \times 10^{-5}$, and 300 warmup steps; and Phi-3-Medium uses a batch size of 4, learning rate of $5 \times 10^{-5}$, and 500 warmup steps.

\subsection{Language Generation Task}\label{sec:train_lg}
For training on language generation tasks, we utilize the training set of the PG19 dataset. We employ the WarmupCosine learning rate scheduler with 500 warmup steps and the AdamW optimizer with $(\beta_1, \beta_2) = (0.9, 0.95)$ and weight decay 0.01 for 1 epoch. The hyperparameters are adjusted for each model: TinyLlama-1.1B uses a batch size of 8 and a learning rate of $5 \times 10^{-5}$; LLaMA-3-8B uses a batch size of 4 and a learning rate of $2 \times 10^{-5}$; and Phi-3-Medium uses a batch size of 2 and a learning rate of $1 \times 10^{-5}$.

\begin{table}[b!]
    \centering
    \begin{minipage}{0.99\columnwidth}
    \vspace{0mm}    
    \centering
    \caption{Templates used for each task in training on language understanding tasks}
    \label{tab:template_training}
    \begin{tcolorbox} 
        \centering
        \fontsize{8pt}{10pt}\selectfont
        \begin{tabular}{p{0.99\columnwidth}}
                \VarSty{BoolQ:}

                \{passage\}\texttt{\textbackslash n}Question: \{question\}\texttt{\textbackslash n}Answer:

                \VarSty{CB:}

                \{premise\}\texttt{\textbackslash n}Question: \{hypothesis\}. True, False, or Neither?\texttt{\textbackslash n}Answer:

                \VarSty{CoPA:}

                \{premise\} so/because \{candidate\}

                \VarSty{SST2:}

                \{sentence\}\texttt{\textbackslash n}Question: Is this sentence positive or negative?\texttt{\textbackslash n}Answer:

                \VarSty{RTE:}
                
                \{premise\}\texttt{\textbackslash n}Question: \{hypothesis\} True or False?\texttt{\textbackslash n}Answer:
                
        \end{tabular}
    \end{tcolorbox}
    \vspace{-3mm}
    \end{minipage}
\end{table}

\section{Evaluation Details}
\subsection{Language Understanding Task}\label{sec:eval_lu}
Unless otherwise specified, we use the task templates introduced in lm-evaluation-harness~\citep{eval-harness} for all our evaluations on language understanding tasks.
We report the accuracy for task BoolQ, CoPA, SST2, CB, RTE, OpenbookQA, ARC-Challenge, Winogrande, PIQA, and ARC-Easy, while report the normalized accuracy for Hellaswag.

For a fair comparison when using multi-shot demonstration contexts, we generate the required number of demonstrations from the training set of each task. 
These same demonstrations are then used as context for evaluating all methods.
This approach eliminates potential variability due to demonstration selection, allowing for a more direct comparison of different methods. 
The results we report are averaged from three independent runs.

\paragraph{TempLoRA:}
We apply LoRA to every linear layer of the base model and directly train it on the given in-context examples. For optimization, we use the AdamW optimizer with a OneCycleLR learning rate scheduler. The rank and $\alpha$ of LoRA are both set to 64 across all models. We use a fixed learning rate of $1 \times 10^{-5}$ and train for 5 epochs.

\paragraph{H$_2$O:}
We retain 20\% of the context, with both the heavy ratio and recent ratio set to 0.1.

\paragraph{SnapKV:}
For SnapKV, we allocate 10\% of the context for the observation window and retain an additional 10\% for inference, leading to a total context retention of 20\%.

\paragraph{\modelname:}
Unless otherwise specified, we convert 80\% of the context into a parameter update, leaving the remaining 20\% of the context unchanged.

\begin{table}[t!]
\centering
\caption{The context length of different demonstration of different tasks using LLaMA-3-8B tokenizer}
\label{tab:data_stat}
\resizebox{\columnwidth}{!}{%
\begin{tabular}{@{}lcccccccc@{}}
\toprule
\multirow{2}{*}{Task Name} & \multicolumn{8}{c}{Context Length}                                         \\ \cmidrule(l){2-9} 
                           & 1-shot & 3-shot & 5-shot & 10-shot & 15-shot & 20-shot & 25-shot & 30-shot \\ \midrule
BoolQ                      & 275    & 475    & 744    & 1218    & 2219    & 2992    & 4190    & 4567    \\
CoPA                       & 18     & 44     & 86     & 145     & 232     & 316     & 374     & 450     \\
SST2                       & 28     & 91     & 121    & 290     & 412     & 470     & 705     & 872     \\
CB                         & 97     & 240    & 550    & 988     & 1533    & 1629    & 2703    & 3304    \\
RTE                        & 42     & 161    & 500    & 958     & 1240    & 1666    & 1984    & 2399    \\
Hellaswag                  & 63     & 298    & 382    & 567     & 1051    & 1470    & 1959    & 2129    \\
Winogrande                 & 25     & 77     & 129    & 231     & 322     & 416     & 527     & 638     \\
OpenbookQA (OBQA)          & 20     & 53     & 76     & 164     & 267     & 344     & 393     & 437     \\
ARC-Chanllege (ARC-C)      & 25     & 154    & 171    & 363     & 570     & 680     & 895     & 998     \\
ARC-Easy (ARC-E)           & 29     & 129    & 194    & 367     & 426     & 802     & 983     & 1296    \\
PIQA                       & 23     & 191    & 257    & 371     & 598     & 817     & 973     & 1028    \\ \bottomrule
\end{tabular}%
}
\end{table}

\subsection{Language Understanding Scaling Analysis}\label{appendix:dvp_setting}
We evaluate different methods under varying numbers of demonstrations on six tasks: BoolQ, RTE, SST2, ARC-Challenge (ARC-C), ARC-Easy (ARC-E), and PIQA. To ensure a fair comparison, we employ a consistent approach across all methods. We first generate a fixed set of demonstrations for each task, which is then used as context for all methods being compared. Our evaluation covers 1, 3, 5, 10, 15, 20, 25, and 30-shot scenarios. 
The reported results are obtained by averaging three different runs, each utilizing a distinct set of generated demonstrations.

We also provide the average context length for each task across different numbers of demonstrations using the LLaMA-3-8B tokenizer in Table~\ref{tab:data_stat}.

\begin{figure}[t!]
\centering
\includegraphics[width=\textwidth]{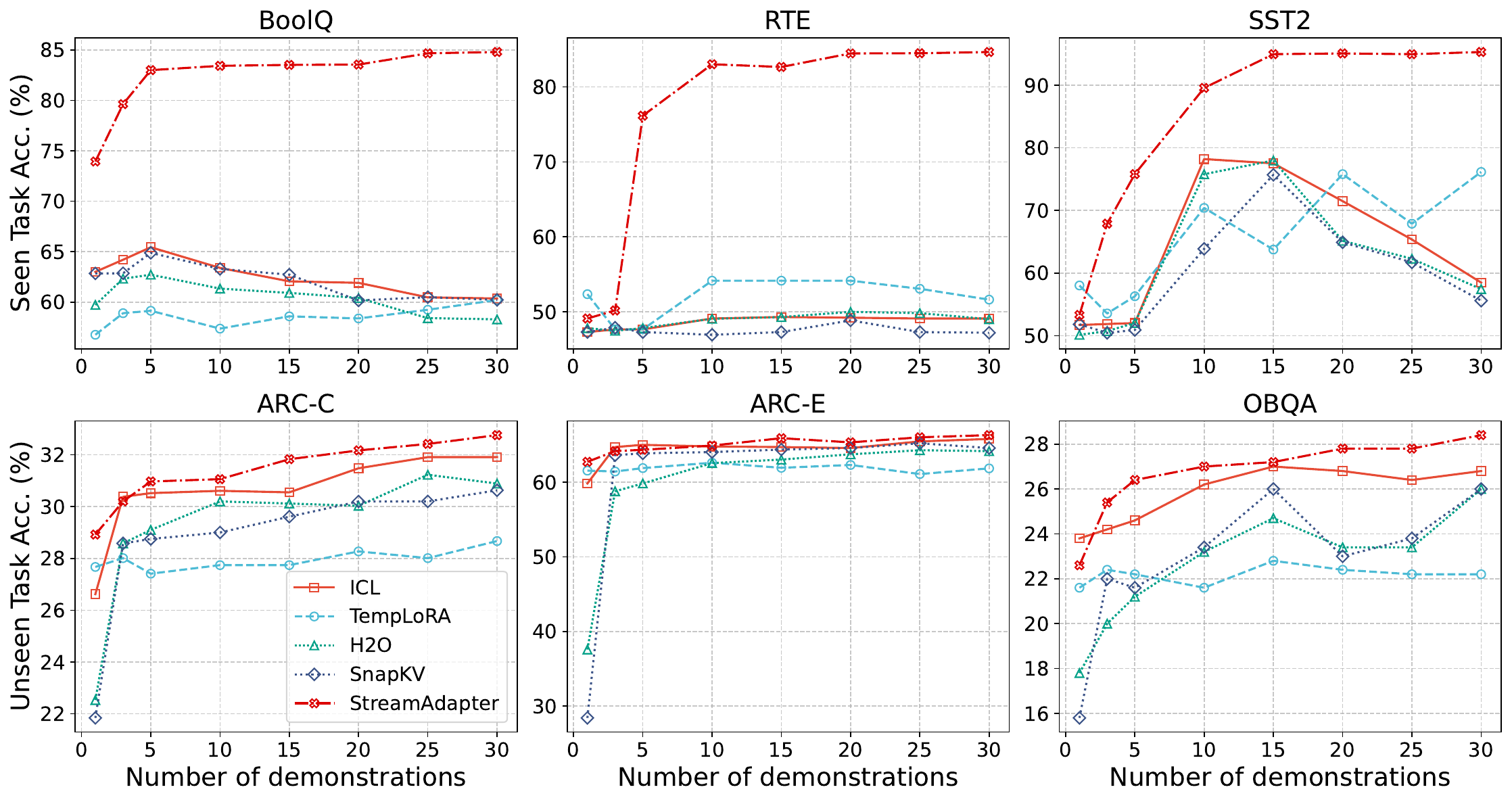}
\caption{Comparison of various methods across different tasks using TinyLlama-1.1B with different numbers of demonstrations}
\label{fig:dvp_tinyllama}
\end{figure}

\begin{figure}[t!]
\centering
\includegraphics[width=\textwidth]{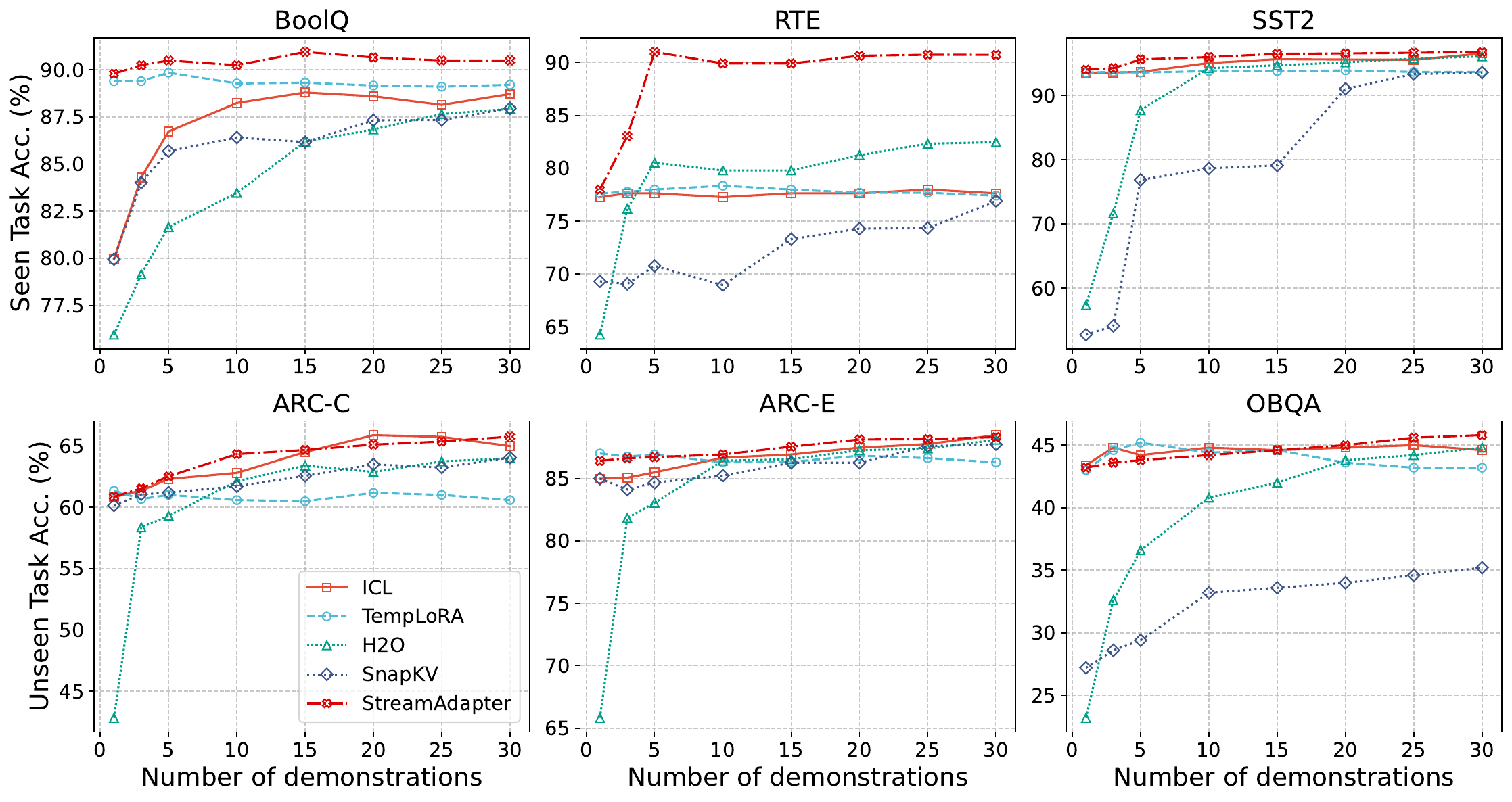}
\caption{Comparison of various methods across different tasks using Phi-3 Medium with different numbers of demonstrations}
\label{fig:dvp_phi3}
\end{figure}

\begin{table}[t!]
\centering
\caption{Average accuracy of \modelname~on language understanding tasks using TinyLlama-1.1B, evaluated across different adaptation ratios}
\label{tab:detail_eval}
\resizebox{\columnwidth}{!}{%
\begin{tabular}{c|cccccc|ccccccc}
\hline
\multirow{2}{*}{Adaptation Ratio} & \multicolumn{6}{c|}{Seen Task}                                                                      & \multicolumn{7}{c}{Unseen Task}                                                                                      \\ \cline{2-14} 
                                       & BoolQ          & CoPA           & SST2           & CB             & RTE            & Avg.           & Hellaswag      & Winogrande     & OBQA           & ARC-C          & ARC-E          & PIQA           & Avg.           \\ \hline
10\%                                   & 70.15          & 76.00          & 80.25          & 55.64          & 51.99          & 66.81          & 59.62          & 60.46          & 26.00          & 30.12          & 65.03          & 73.72          & 52.49          \\
20\%                                   & 74.28          & 75.00          & 83.37          & 55.00          & 76.91          & 72.91          & 59.29          & \textbf{60.69} & 25.80          & 30.38          & 64.86          & 73.88          & 52.48          \\
30\%                                   & 77.83          & \textbf{77.00} & 84.17          & 62.86          & 81.59          & 76.69          & 59.39          & 60.38          & 26.00          & 30.38          & 64.90          & 74.05          & 52.52          \\
40\%                                   & 79.60          & 76.00          & 85.08          & 66.07          & 83.75          & 78.10          & \textbf{59.83} & 60.06          & 25.80          & 30.38          & 65.19          & 73.94          & 52.53          \\
50\%                                   & 80.06          & 75.00          & 83.14          & 81.07          & \textbf{83.75} & 80.60          & 59.28          & 59.80          & 26.00          & 30.55          & \textbf{65.56} & 74.05          & 52.54          \\
60\%                                   & 80.37          & 74.00          & 84.63          & 82.86          & \textbf{83.75} & 81.12          & 59.30          & 60.22          & 26.40          & 30.55          & 65.45          & \textbf{74.16} & \textbf{52.68} \\
70\%                                   & 81.47          & 75.00          & 89.79          & 85.77          & \textbf{83.75} & 83.16          & 59.67          & 59.35          & \textbf{27.00} & \textbf{31.06} & 65.07          & 73.39          & 52.59          \\
80\%                                   & 81.77          & 76.00          & 89.56          & 85.71          & 82.67          & 83.14          & 59.45          & 59.91          & \textbf{27.00} & \textbf{31.06} & 64.93          & 73.29          & 52.61          \\
90\%                                   & 82.36          & 75.00          & 90.31          & 85.72          & 82.67          & 83.21          & 59.54          & 59.04          & 23.00          & 29.60          & 64.10          & 73.18          & 51.41          \\
100\%                                  & \textbf{83.36} & 74.00          & \textbf{93.81} & \textbf{90.43} & 82.67          & \textbf{84.85} & 59.40          & 59.04          & 23.00          & 29.52          & 64.02          & 73.78          & 51.46          \\ \hline
\end{tabular}%
}
\end{table}

\subsection{Language Generation Task}\label{sec:eval_lg}
For TempLoRA, we apply the LoRA adapter to every linear layer, with the rank and $\alpha$ both set to 64.

\subsection{Robustness Analysis}
For evaluting the robustness of ICL and \modelname~adaption capability from different prompt template, we use different prompt template for in-contetx examples and targer sample.
For in-context examples, we use the same prompt teamplte with lm-evaluation-harness~\citep{eval-harness}, while the teamplte for the target sample are presented in Table~\ref{tab:template_robutness}.

\begin{table}[t!]
    \centering
    \begin{minipage}{0.99\columnwidth}
    \vspace{0mm}    
    \centering
    \caption{Templates used for each task in training on language understanding tasks}
    \label{tab:template_robutness}
    \begin{tcolorbox} 
        \centering
        \fontsize{8pt}{10pt}\selectfont
        \begin{tabular}{p{0.99\columnwidth}}
                \VarSty{BoolQ:}

                Given the passage: \{passage\}\texttt{\textbackslash n}Answer the following question: \{question\}?

                \VarSty{SST2:}

                Could you tell me is this <\{sentence\}> positive or negative?

                \VarSty{RTE:}
                
                \{premise\}\texttt{\textbackslash n}\{hypothesis\} True or False?

                \VarSty{ARC-Challenge:}

                \{question\}\texttt{\textbackslash n}Please answer the question.

                \VarSty{ARC-Easy:}

                \{question\}\texttt{\textbackslash n}Please answer the question.

                \VarSty{PIQA:}

                Answer the question: \{question\}.
        \end{tabular}
    \end{tcolorbox}
    \vspace{-3mm}
    \end{minipage}
\end{table}

\section{Additional Results}
\subsection{Scaling Analysis on Language Understanding Task}
We further present the results on language understanding tasks with varying numbers of demonstrations for TinyLlama-1.1B and Phi-3-Medium in Figure~\ref{fig:dvp_tinyllama} and Figure~\ref{fig:dvp_phi3}, respectively.
These results further demonstrate that \modelname~clearly outperforms full ICL and other TTA methods. Moreover, \modelname~exhibits better scaling capability as the number of demonstrations increases.

\subsection{Evaluation with Different Adaptation Ratio}\label{appendix:detail_adaptation}
Table~\ref{tab:detail_eval} presents the detailed accuracy of \modelname across different adaptation ratios, as discussed in Section~\ref{sec:analysis}.

\end{document}